 \documentclass[tablecaption=bottom,pmlr]{jmlr}

\usepackage{booktabs}
\usepackage[load-configurations=version-1]{siunitx} 

 \usepackage{xurl}

\theorembodyfont{\upshape}
\theoremheaderfont{\scshape}
\theorempostheader{:}
\theoremsep{\newline}

\jmlrvolume{1}
\jmlryear{2021}
\jmlrsubmitted{submission date}
\jmlrpublished{publication date}
\jmlrworkshop{NeurIPS 2020 Preregistration Workshop} 

\title{Latent Neural Differential Equations\;\\ for Video Generation}

  \author{\Name{Cade Gordon} \Email{cadegordonml@gmail.com}\and
   \Name{Natalie Parde} \Email{parde@uic.edu}\\
   \addr Department of Computer Science \\ University of Illinois at Chicago}

\begin{document}

\maketitle

\begin{abstract}
  Generative Adversarial Networks have recently shown promise for video generation, building off of the success of image generation while also addressing a new challenge: time. Although time was analyzed in some early work, the literature has not adequately grown with temporal modeling developments. We study the effects of Neural Differential Equations to model the temporal dynamics of video generation. The paradigm of Neural Differential Equations presents many theoretical strengths including the first continuous representation of time within video generation. In order to address the effects of Neural Differential Equations, we investigate how changes in temporal models affect generated video quality. Our results give support to the usage of Neural Differential Equations as a simple replacement for older temporal generators. While keeping run times similar and decreasing parameter count, we produce a new state-of-the-art model in 64$\times$64 pixel unconditional video generation, with an Inception Score of 15.20.
\end{abstract}

\section{Introduction}

Generative modeling remains an important problem within computer vision, with new developments providing a better understanding of high-dimensional data modeling and even aiding the supervised learning sphere. Good representations of distributions improve
feature space visualisation, clustering, and classification.  Many approaches have tackled the problem of representing a distribution, including Generative Adversarial Networks (GANs) \citep{gans}, which have recently shown immense potential for image generation. As time progresses GANs become more robust, allowing for greater image size \citep{dcgan,proggan,biggan} and quality \citep{stylegan1,stylegan2}.

The success of GANs in image generation propelled them towards being the prominent methodology for video generation. However, the application of GANs to video generation has come with new challenges. Adding time to the preexisting color, width, and height dimensions has increased computational costs and complexity by an order of magnitude. Early models generated videos of a meagerly 64 by 64 pixels \citep{vgan,tgan,mocogan}. The addition of the new temporal component not only restricted video size, it also opened many questions regarding the best way to navigate an entirely new dimension. The first model to use GANs for video generation was VGAN \citep{vgan}, which used 3D convolutional kernels to account for time, framing it as no more than an extra feature channel blended in with color, width and height.

Treating temporal features as a separate dimensional scope allowed for the subsequent TGAN \citep{tgan} to outperform VGAN in terms of Inception Score (IS) \citep{inceptionscore}. The authors proposed two separate generative architectures: a 1D convolutional temporal generator and an image generator. Further investigation of the temporal latent space was done by MoCoGAN \citep{mocogan}, in which the authors proposed decomposing the image generator's input into a single content vector and an evolving motion vector. Experimentation has also gone into increasing frame size and network depth, with some reflections on computational mitigation \citep{tganv2,dvdgan,moflowgan}, but little work has gone into rigorously examining time.

Our work reopens the discussion of the temporal latent space. After the revelation of separate temporal generation, researchers have stopped asking questions about the temporal  generator. Works after TGAN employed Long Short-Term Memory (LSTM) \citep{lstm} or Convolutional LSTM (CLSTM) \citep{clstm} blocks. To this day, the LSTM remains and has never been fully ablated or examined with control. A similar previously unproven but accepted notion was content motion decomposition. First published in 2018 as part of MoCoGAN \citep{mocogan}, content and motion decomposition was not ablated until the publication of MoFlowGAN \citep{moflowgan} in 2020. With very limited analysis, much of the temporal space remains an open question within these models.

While able to model temporal dynamics, recurrent models such as the LSTM and its variants only represent discrete samples. We propose to re-explore the temporal space under a continuous paradigm. Neural Ordinary Differential Equations (NODEs) \citep{neuralode} offer the potential for a continuous representation of the temporal dimension. Extending the paradigm of Neural Differential Equations, we propose the first continuous video generation model.  Our work makes the following contributions:

\begin{itemize}
    \item We establish the first continuous GAN for video generation.
    \item We experiment with multiple novel architectures for video generation.
    \item We analyze how changes in the temporal latent space modality affect visual fidelity through an ablation study.
\end{itemize}


\section{Related Work}

\subsection{Generative Adversarial Networks}

Two neural networks compose GANs: a Discriminator $D$ and Generator $G$. The generator transforms a sampled noise vector $z$ from a distribution $p_z$ and maps it to an image (or in our case a video). The Discriminator functions by taking an input image or video $x$ and mapping it to a value representing whether it believes $x$ is sampled from the real distribution $p_x$ or the distribution produced by the generator $p_g$. The two compete to minimize or maximize a loss function that may be represented generically as shown below, where $\phi$ is a function of the Discriminator's prediction and the truth label represented as $1$ (real) or $0$ (fake):

$$
\max_G \min_D \mathbb{E}_{x\sim p_x}[\phi(D(x),1)] + \mathbb{E}_{z\sim p_z}[\phi(D(G(z)),0)]
$$

$\phi$ is typically the identity function, cross entropy function, or hinge loss function. Loss choice has been shown to be less consequential so long as a Lipschitz constraint is met \citep{lossfuncgan}. GANs are often difficult to train, and two approaches to increasing their stability during training time are through applying a form of Lipschitz constraint or multi-scale generation. WGAN \citep{wgan} and WGAN-GP \citep{wgangp} showed the effectiveness of the Lipschitz regularization. SNGAN \citep{sngan} subsequently showed a refined way to enforce the constraint through spectral normalization. Progressive GAN \citep{proggan} stabilized training by increasing the generated resolution over time.

\subsection{Video Prediction}

Video prediction conditions a model on a sample of frames, and models the subsequent frames. A common approach is the use of a recurrent architecture such as an LSTM \citep{vidrecurrent1,vidrecurrent2,vidrecurrent3,vidrecurrent4,vidrecurrent5,vidrecurrent6}. Another common methodology is using optical flow \citep{flow1,flow2,flow3,flow4}. Prior work has also explored the stochastic nature of videos \citep{stochasticgen,stochasticvar,savp,slatentres,stochhighfid}.

\subsection{Video Generation}

\begin{figure}
    \centering
    \includegraphics[scale=0.5]{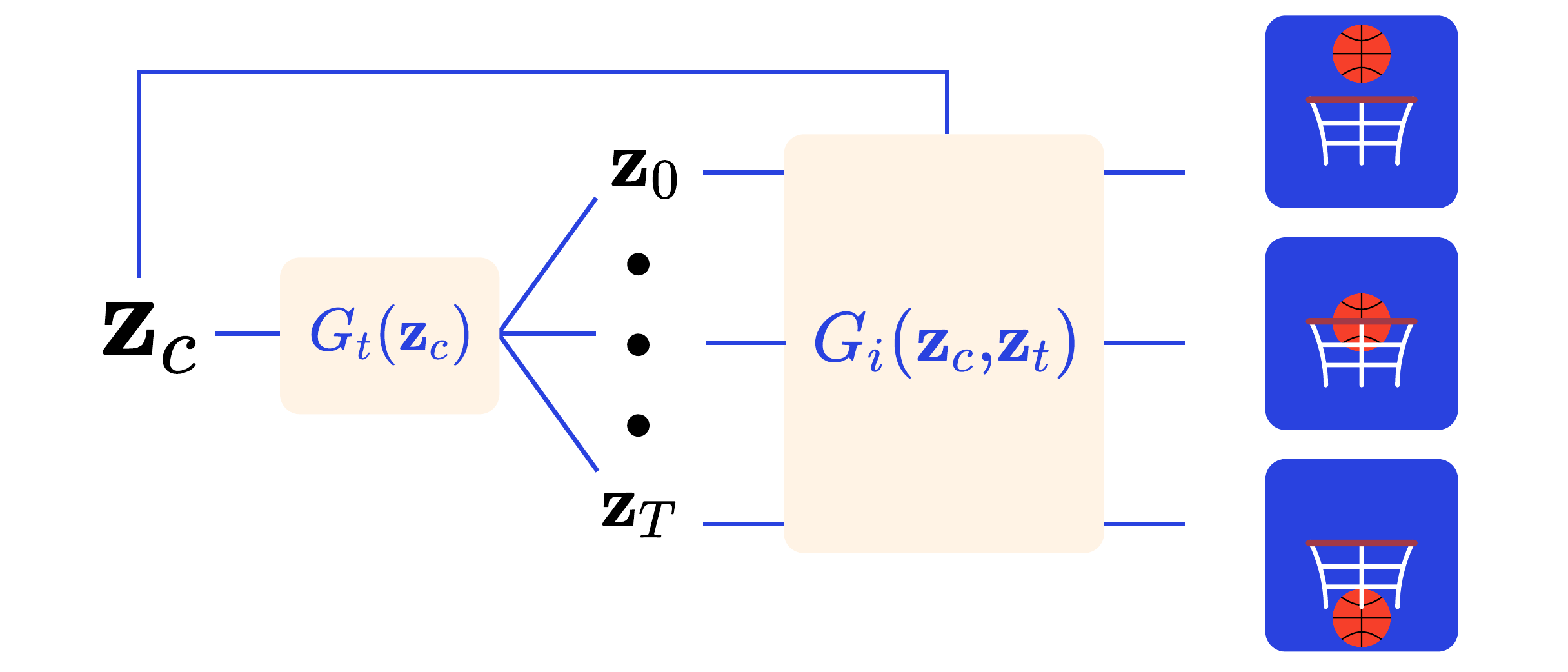}
    \caption{A latent variable $\mathbf{z}_c$ is transformed by $G_t$ into a series of temporal vectors $\mathbf{z}_0, \mathbf{z}_1, ..., \mathbf{z}_T$. Each temporal vector $\mathbf{z}_t$ is concatenated with $\mathbf{z}_c$ and transformed into an image. Said images are joined to compose a video.}
    \label{fig:vid_gen}
\end{figure}

To the best of our knowledge, the first work to use a GAN to generate videos was VGAN \citep{vgan}. VGAN generated videos using  spatio-temporal convolutions with 3D kernels and fractional strides, separately generating the motion and background. In order to combine the two it used a learned mask to produce the final output. 

Its successor, TGAN \citep{tgan}, separately generated temporal and frame features. TGAN transformed a single noise vector into multiple vectors accounting for time with a temporal generator $G_t$, a series of 1D convolutions. The generated vectors concatenated with the starting single noise vector were then fed into an image generator $G_i$. By separating temporal generation into its own process, TGAN outperformed VGAN \citep{inceptionscore}. The general form of $G$ may be seen in Figure \ref{fig:vid_gen}.

MoCoGAN \citep{mocogan} continued in the line of temporal manipulation by using an LSTM to generate temporal features. The authors assumed that the temporal space was composed of a motion and content subspace. Though their work outperformed TGAN, it was not until MoFlowGAN \citep{moflowgan}, two years later, that the content and motion decomposition was fairly ablated, with results showing that it led to a positive increase in IS. It is hard to know for many of these models which features actually allowed for their success, since the discriminator and image generation architectures change and increase in complexity from one paper to the next. In light of this, a properly controlled analysis of our proposed model will fill in many of the gaps in the current literature.

Other papers focus on increasing the dimension of the video output. DVD-GAN and MoFlowGAN \citep{dvdgan, moflowgan} produce 128x128 pixel videos. The current state-of-the-art TGANv2 \citep{tganv2} even boasts 192x192 pixel videos. Recently TGAN-F \citep{tganf} further improved performance by simplifying the discriminator of TGAN.

\subsection{Neural Differential Equations}

NODEs \citep{neuralode} transformed the vision of ResNets \citep{resnet} by giving them a continuous definition. Instead of the singular discrete additions of a neural network function $f(x)$, they proposed integrating using ordinary differential equation (ODE) solvers.  The new interpretation allows for an approximate continuous temporal representation, where $\mathbf{h}$ represents the hidden state of a layer, and $t$ represents the ordering of layers:

$$\mathbf{h}_{t+1} = \mathbf{h}_t + f(\mathbf{h}_t) = \mathbf{h}_t + \int_t^{t+1} g(\mathbf{h}_t,t)t$$
$$\frac{d\mathbf{h}_t}{dt}=g(\mathbf{h}_t,t)$$

We use $t$ to represent time.
Works like $\textrm{ODE}^2\textrm{VAE}$ \citep{ode2vae} extended this to second order ODEs. Much of the work surrounding NODEs revolves around Variational Autoencoders \citep{neuralode, ffjord, ode2vae}. Recently, even more differential equation families have been explored; Neural Stochastic Differential Equations (NSDEs) are a successful example \citep{sde, sde2}.

\section{Neural Differential Equation Video GANs}

\begin{figure}
    \centering
    \includegraphics[scale=0.5]{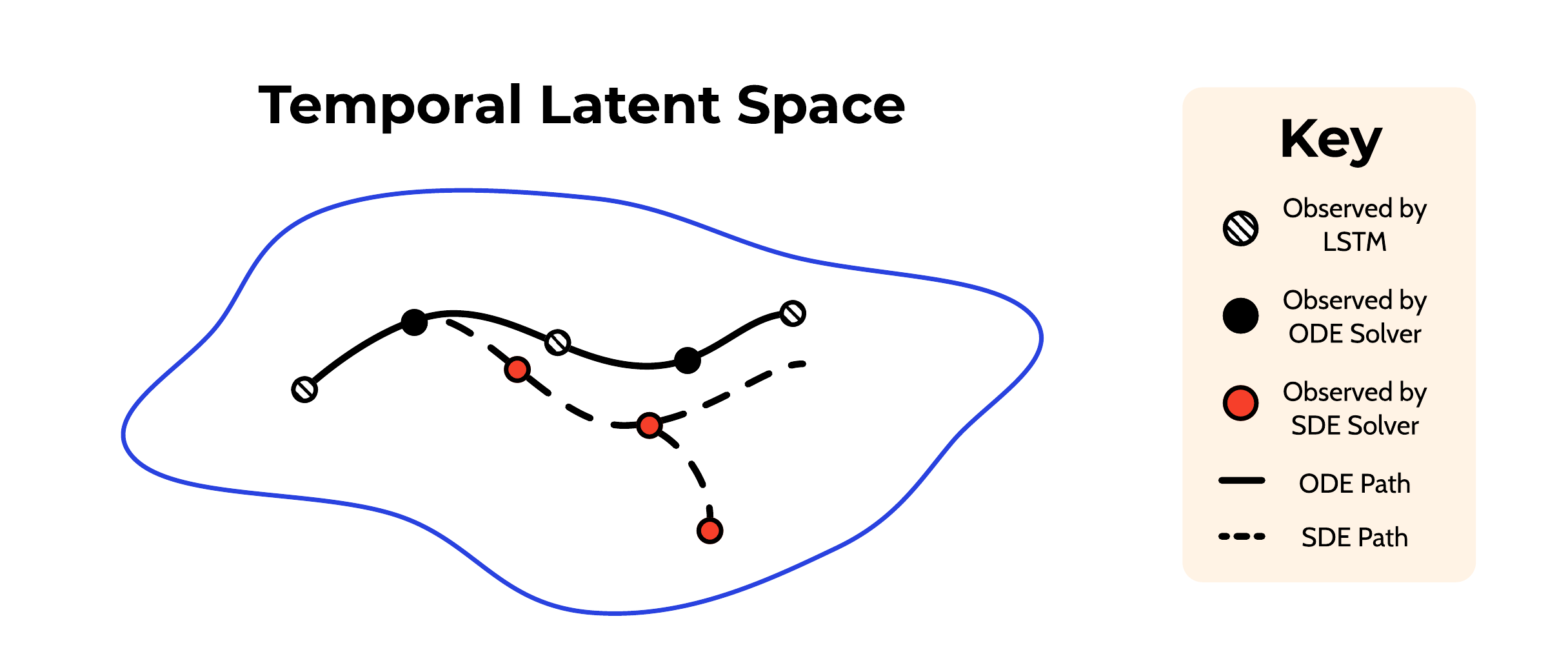}
    \caption{When compared with typical LSTMs, neural differential equations have more frequent observations, and SDEs have greater potentiality for solutions.}
    \label{fig:latent_space}
\end{figure}

There is a significant gap in the literature explaining the choice for temporal generators. To remedy this, we propose to explore it under a paradigm common to general physics: using differential equations to represent temporal dynamics. While using historical image generator functions, we will observe changes in performance metrics with different temporal generator functions. Comparisons between the families of generative functions may be visualized by Figure \ref{fig:latent_space}.

\subsection{Ordinary Differential Equations (ODEs)}

Instead of an auto-regressive LSTM or 1D Kernel, a differential equation may be used to model the evolution of a latent variable $\mathbf{z}_t$. By a learned function $f(\mathbf{z}_t)$, future $\mathbf{z}_{T}$ may be found by integrating:

$$\mathbf{z}_{T} = \mathbf{z}_0 + \int_0^{T} f(\mathbf{z}_t,t) dt$$
$$\dot{\mathbf{z}_t}=\frac{d\mathbf{z}_t}{dt}=f(\mathbf{z}_t,t)$$

The image generator $G_i(\mathbf{z})$ may then produce an image from $\mathbf{z}_t$. Using a differential equation the model may account for the finer nuances of traversing the latent space accounting for motion in $\mathbf{z}_{t<t+\epsilon<t+1}$. LSTMs only view sparse time steps; the model moves from $\mathbf{z}_t$ to $\mathbf{z}_{t+1}$ never accounting for a $\mathbf{z}_{t+0.5}$. NODEs allow for the intermediate $\mathbf{z}_t$ values to be traversed, which may potentially lead to better performance as this can more closely approximate a latent trajectory.

The family of NODEs also allows for higher order interpretations of the model. Our $f(\mathbf{z}_t)$ may represent higher orders than simple $\dot{\mathbf{z}}_t$, such as $\ddot{\mathbf{z}}_t$ or higher. A first-order ODE parameterizes more immediate changes during integration, whereas higher orders represent much more long-term shifts, such as concavity, in the latent variable.

\subsection{Stochastic Differential Equations (SDEs)}

NODEs allow for path approximations in determinate systems. Every $\mathbf{z}_t$ will produce a single $\mathbf{z}_{t+1}$, but this isn't reflective of how videos truly function. There is an inherent stochastic nature to how videos progress---actors have a branching tree of decisions and so do particles for their motion. NSDEs may be a good way to represent the random nature present in videos, offering all of the benefits of NODEs while allowing for randomness with their added noise. Under this form, and letting $\mu(\mathbf{z}_t)$ and $\sigma(\mathbf{z}_t)$ represent drift and diffusion respectively, we find $\mathbf{z}_{T}$ with:

$$\mathbf{z}_{T} = \mathbf{z}_0 + \int_0^T \mu(\mathbf{z}_t,t) dt + \int_0^{T} \sigma(\mathbf{z}_t,t) dW_t$$

Each of $\mu(\mathbf{z}_t)$ and $\sigma(\mathbf{z}_t)$ are parameterized by a neural network. $W_t$ is a Wiener process, a continuous series of values with Gaussian increments.  The validity of this formulation may be exemplified by thinking about a video of a face changing expressions. If the actor starts out with a neutral face they may then produce a sad one after that. However, a smile would be equally likely. By injecting randomness either path may be explored by the model.

\subsection{Benefits of Differential Equations}

Differential equations allow for increased control over how paths are traversed because of their continuous properties. Because $\mathbf{z}_t$ is found by integration, there are two unique characteristics that other modalities do not possess. First, $\mathbf{z}_t$ can be integrated backwards in time allowing the discovery of $\mathbf{z}_{t-n}$. This can be thought of as what happens before the first frame. Second, if increased frame rates are desired, they can easily be accounted for. The differential equation solver will necessitate evaluations of $\mathbf{z}_{t<t+\epsilon<t+1}$. To achieve a higher frame rate, the image generator simply needs to sample some of the intermediate $\mathbf{z}_t$ evaluations. Control like this is impossible in recurrent models.

\section{Experimental Protocol}

The most widely used and comparable metrics for video generation are IS and Fr\'echet Inception Distance (FID) \citep{ttur}. These are calculated by a C3D model \citep{c3d} pretrained on the UCF101 dataset \citep{ucf101}.
Their values quantify visual fidelity of the generated videos. We observe changes to these metrics as we alter $G_t$. We train the following models on UCF101:
\begin{itemize}
    \item TGAN
    \item MoCoGAN
    \item TGANv2 (used for Effects of Family and Order only)
\end{itemize}

Each model will run for 100,000 epochs using the model's originally proposed hyperparameters. IS will be calculated on samples of 2,048 videos every 2,000 epochs. The epoch with the highest IS will be used to calculate the model's final statistics. Using the best performing epoch, five batches of 2,048 videos will be created. FID and IS will be calculated on each batch, and we will report the mean value and standard deviation for each metric.

\subsection[Finding f(x)]{Finding $f(x)$}

In order to generate videos under this paradigm an appropriate neural network architecture for $f(x)$ needs to be studied. To find $f(x)$, TGAN and MoCoGAN will be trained under different $G_t$s. For each, $\mathbf{z}_t$ will be found by integrating $f(x)$ as a first order NODE. Ablation will occur with the following $f(x)$s: $f:\mathbb{R}^d\to\mathbb{R}^d$ using a single learned layer with a nonlinearity; $f,g,h:\mathbb{R}^d\to\mathbb{R}^d$ where $f(x)=(g \circ h)(x)$, with $g$
and $h$ being also single learned layers with a nonlinearity; and the same functions as the previous setup but with $g$ and $h$ equalizing parameters of each model's original $G_t$. Testing these choices of $f(x)$
across both TGAN and MoCoGAN allows for greater evidence for or against how well each $f(x)$ generalizes to the task and architecture.

\subsection{Effects of Family and Order}

With an effective $f(x)$, we can ablate the multiple families and orders. TGAN, MoCoGAN, and now TGANv2 will be tested under the following motion generators: the model's original $G_t$, the first order ODE, the second order ODE, the third order ODE, and the SDE. For each configuration we will report IS and FID using the process specified earlier.

\section{Implementation}

In this section we further detail model architectures and explain minor alterations to the planned experimental design. All experiments are performed on an NVIDIA RTX 2080 TI GPU and are written in PyTorch. To promote future research and replication we make our code available to the community.\footnote{\url{https://github.com/Zasder3/Latent-Neural-Differential-Equations-for-Video-Generation}}

\subsection{Further Specification}

As outlined above, said $f(x)$ designs necessitated careful measures to make them functional and fairly comparable to their original model's counterparts. In all models except TGANv2, $\tanh$ composes the final or intermediate activation function. It satisfies the continuous and Lipschitz constraint for uniqueness specified by \citet{neuralode}. Furthermore, as the final layer it permits both positive and negative resulting values stopping $\mathbf{z}_t$ from being monotonically increasing with time.

Additionally, we feed the starting noise vector through a fully connected network (FCN) aiming to equalize the number of nonlinearities. This design choice originated from comparing the number of activation functions in $G_t$ in our experimental groups to those of the original models. For example, in TGAN the temporal generator is composed of four ReLUs and a final $\tanh$. As it stands, $f(x)$ has only one nonlinearity. To amend the nonlinearity gap between our proposed $G_t$ and the original models, we prepended an FCN to $f(x)$ with equal nonlinearity count to the original model's $G_t$. This means when integrating, individual $z_t$s will be in a comparatively complex space. We follow this protocol of prepending the FCN to the integration for all experiments except those with TGANv2 and MoCoGAN with equal parameterization. 

\subsection{Deviations from Original Plans}

Instead of calculating IS every 2,000 training iterations, we calculated IS every 1,000 iterations. We also increased the number of samples used to compute an IS and FID mean and standard deviations. The original 5 measures became 10 to increase precision and to become more inline with that of TGANv2's protocol. 

We also found it infeasible to compute every $f(x)$ family variation of TGANv2 under our setup due to limited computational resources. Opting to train only a first order ODE, with a batch size of 32, the model took three days to train on an A100 GPU costing over \$350 using Google Cloud Platform. 

\section{Results}

\subsection[Finding F(x)]{Finding $f(x)$}
\label{sec:finding_fx}

\begin{table}[hbtp]
\centering
\begin{tabular}{lllll}
$f(x)$ Type & Original & Single Layer & Two Layers & Equal Parameters \\ \toprule
TGAN        &15.06$\pm$0.25&\textbf{15.20$\pm$0.26}&14.39$\pm$0.27&14.08$\pm$0.18\\ \midrule
MoCoGAN     &10.86$\pm$0.16&10.24$\pm$0.16&9.70$\pm$0.14&\textbf{12.61$\pm$0.21} \\ \bottomrule
\end{tabular}
\caption{Inception Score by type of $f(x)$ (higher is better)}
\label{tab:func_is}
\end{table}

\begin{table}[hbtp]
\centering
\begin{tabular}{lllll}
$f(x)$ Type & Original & Single Layer & Two Layers & Equal Parameters \\ \toprule
TGAN        &\textbf{26512}$\pm$27&26678$\pm$21&26750$\pm$21&26751$\pm$27 \\ \midrule
MoCoGAN     &27951$\pm$28&28767$\pm$61&28967$\pm$41&\textbf{26998$\pm$33}\\ \bottomrule
\end{tabular}
\caption{Fr\'echet Inception Distance by type of $f(x)$ (lower is better)}
\label{tab:func_fid}
\end{table}

Looking to the IS and FID across different variations of $f(x)$ in Tables \ref{tab:func_is} and \ref{tab:func_fid}, we find a loose trend relating performance of the model to parameter count. Within the TGAN runs, parameters increase from left to right, but the IS decreases from left to right. In the case of MoCoGAN, equal parameters actually significantly increase performance in comparison to the single layer and two layer models as this variation forced the removal of the embedding FCN. More research needs to be done to conclude this hypothesis, but as it stands parameter count has predictive power on model performance.

The previous state-of-the-art IS for unconditional 64$\times$64 pixel on UCF101 was held by TGAN-F, with an average IS of 13.62. Our variant which we will term TGAN-ODE outperforms this mark to become the new state-of-the-art, with an average IS of 15.20.

\subsection{Effects of Family and Order}

\begin{table}[hbtp]
\centering
\begin{tabular}{llllll}
Family  & Original & 1st Order & 2nd Order & 3rd Order & SDE \\ \toprule
TGAN    &15.06$\pm$0.25&\textbf{15.20$\pm$0.26}&13.96$\pm$0.23&13.39$\pm$0.20&14.62$\pm$0.28\\ \midrule
MoCoGAN &10.86$\pm$0.16&12.61$\pm$0.21&11.84$\pm$0.22&11.16$\pm$0.18&\textbf{14.33$\pm$0.21}\\ \midrule
TGANv2  &\textbf{26.60$\pm$0.47}\footnotemark[2]  &21.02$\pm$0.28&     -          &        -      & -   \\ \bottomrule
\end{tabular}
\caption{Inception Score by Family and Order (higher is better)}
\label{tab:fam_is}
\end{table}

\begin{table}[hbtp]
\centering
\begin{tabular}{llllll}
Family  & Original & 1st Order & 2nd Order & 3rd Order & SDE \\ \toprule
TGAN    &\textbf{26512$\pm$27}&26678$\pm$21&26963$\pm$26&27223$\pm$23&27252$\pm$11\\ \midrule
MoCoGAN &27951$\pm$28&\textbf{26998$\pm$33}&27889$\pm$47&28164$\pm$25&28064$\pm$33\\ \midrule
TGANv2  &\textbf{3431$\pm$19}\footnotemark[2]&26017$\pm$29&     -          &        -      & -   \\ \bottomrule
\end{tabular}
\caption{Fr\'echet Inception Distance by Family and Order (lower is better)}
\label{tab:fam_fid}
\end{table}

\footnotetext[2]{Value sourced from original paper instead of reproduced.}

By nature, these experiments were more exploratory than those in \S{}\ref{sec:finding_fx}; however, they produced some noteworthy anomalies and trends. Within our setup we found that performance degrades with increasing order of the ODE across both TGAN and MoCoGAN. The most surprising result is that of MoCoGAN-SDE, which outperformed the baseline and first order implementation by a large margin.

Our entries for TGANv2's original scores are sourced from the paper. Differences in data pipelines, framework, and implementation makes the direct comparison imperfect, but a good proxy for current results. Discrepancies are most noticeable in FID because we did not have access to the original dataset statistics, hence we had to calculate our own. Further work must be done to provide a thorough outcome, but as it stands a first order ODE performs adequately on a large scale, albeit not yet competitively.

\section{Discussion}

From our analysis on small scale models, we find promising results in the usage of ODEs and SDEs as drop-in replacements for the typical temporal generator. Within our experiments we achieve success at parameter counts equal to or lesser than baseline models. Run times also remained nearly identical to the original models. Differential equations seem to provide theoretical and quantitative boosts without harming speed. We find promising evidence of differential equation success at smaller scales, but not yet at larger ones. This opens room for future researchers to more thoroughly investigate scaling the presented technique.

In order to achieve success with these models in higher dimensions, several considerations are necessary. First, with respect to the actual $f(x)$ to be integrated, although we found a suitable function in our quite small search space, it's evident that the choice in function can have drastic effects on our results. Second, larger models come with increased VRAM necessities. A single consumer GPU will no longer be able to handle the current models at scale.

Although not strictly related to our questions of interest, during our training we additionally noted a troubling phenomenon with regard to IS score. From one calculation to the next there was extreme variation in the observed value---at one point in time the model weights may produce that of state-of-the-art, and the next nowhere close. Under older measurement frameworks (for example, only calculating IS on the training end) true model improvements may have been missed. On the other hand, this may have confounded success in models with no true advantage, but rather more luck on the final IS evaluation.

\begin{figure}
    \centering
    \includegraphics[scale=0.5]{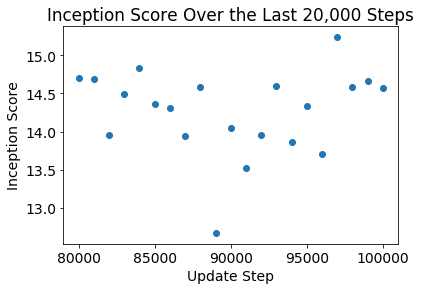}
    \caption{The last 21 calculated IS values from TGAN trained with a first order ODE. Observations range from above 15 (state-of-the-art) to below 13 showing how without careful observation great models might be missed.}
    \label{fig:is_values}
\end{figure}

\section{Conclusion}

Our work presents the first continuous time GAN for video generation and seeks to reopen the question of temporal generation. We find evidence supporting the use of differential equations as potential drop-in replacements for common temporal generators.  We ablate under different integrated functions, differential equation orders, and families to investigate the robustness of differential equations in video generation. On the UCF101 dataset, our variant, termed TGAN-ODE, presents a new state-of-the-art on unconditional 64$\times$64 pixel image generation.

The results of this work reopen the case for investigating the temporal generator and provide a novel direction for others to build upon. We are eager to see the outcomes of researchers' efforts as they scale the video size, use the models under different problem formulations, and increase the frame-rate to further explore this paradigm.

\bibliography{refs.bib}

\begin{thebibliography}{44}
\providecommand{\natexlab}[1]{#1}
\providecommand{\url}[1]{\texttt{#1}}
\expandafter\ifx\csname urlstyle\endcsname\relax
  \providecommand{\doi}[1]{doi: #1}\else
  \providecommand{\doi}{doi: \begingroup \urlstyle{rm}\Url}\fi

\bibitem[Arjovsky et~al.(2017)Arjovsky, Chintala, and Bottou]{wgan}
Martin Arjovsky, Soumith Chintala, and L{\'e}on Bottou.
\newblock Wasserstein generative adversarial networks.
\newblock In \emph{Proceedings of the 34th International Conference on Machine
  Learning-Volume 70}, pages 214--223, 2017.

\bibitem[Babaeizadeh et~al.(2018)Babaeizadeh, Finn, Erhan, Campbell, and
  Levine]{stochasticvar}
Mohammad Babaeizadeh, Chelsea Finn, Dumitru Erhan, Roy~H Campbell, and Sergey
  Levine.
\newblock Stochastic variational video prediction.
\newblock In \emph{International Conference on Learning Representations}, 2018.

\bibitem[Brock et~al.(2018)Brock, Donahue, and Simonyan]{biggan}
Andrew Brock, Jeff Donahue, and Karen Simonyan.
\newblock Large scale gan training for high fidelity natural image synthesis.
\newblock In \emph{International Conference on Learning Representations}, 2018.

\bibitem[Byeon et~al.(2018)Byeon, Wang, Kumar~Srivastava, and
  Koumoutsakos]{vidrecurrent5}
Wonmin Byeon, Qin Wang, Rupesh Kumar~Srivastava, and Petros Koumoutsakos.
\newblock Contextvp: Fully context-aware video prediction.
\newblock In \emph{Proceedings of the European Conference on Computer Vision
  (ECCV)}, pages 753--769, 2018.

\bibitem[Chen et~al.(2018)Chen, Rubanova, Bettencourt, and Duvenaud]{neuralode}
Ricky T.~Q. Chen, Yulia Rubanova, Jesse Bettencourt, and David~K Duvenaud.
\newblock Neural ordinary differential equations.
\newblock In S.~Bengio, H.~Wallach, H.~Larochelle, K.~Grauman, N.~Cesa-Bianchi,
  and R.~Garnett, editors, \emph{Advances in Neural Information Processing
  Systems 31}, pages 6571--6583. Curran Associates, Inc., 2018.
\newblock URL
  \url{http://papers.nips.cc/paper/7892-neural-ordinary-differential-equations.pdf}.

\bibitem[Clark et~al.(2019)Clark, Donahue, and Simonyan]{dvdgan}
Aidan Clark, Jeff Donahue, and Karen Simonyan.
\newblock Adversarial video generation on complex datasets.
\newblock \emph{arXiv}, pages arXiv--1907, 2019.

\bibitem[Denton and Fergus(2018)]{stochasticgen}
Emily Denton and Rob Fergus.
\newblock Stochastic video generation with a learned prior.
\newblock In \emph{International Conference on Machine Learning}, pages
  1174--1183, 2018.

\bibitem[Finn et~al.(2016)Finn, Goodfellow, and Levine]{vidrecurrent3}
Chelsea Finn, Ian Goodfellow, and Sergey Levine.
\newblock Unsupervised learning for physical interaction through video
  prediction.
\newblock In \emph{Advances in neural information processing systems}, pages
  64--72, 2016.

\bibitem[Franceschi et~al.(2020)Franceschi, Delasalles, Chen, Lamprier, and
  Gallinari]{slatentres}
Jean-Yves Franceschi, Edouard Delasalles, Mickaël Chen, Sylvain Lamprier, and
  Patrick Gallinari.
\newblock Stochastic latent residual video prediction, 2020.

\bibitem[Goodfellow et~al.(2014)Goodfellow, Pouget-Abadie, Mirza, Xu,
  Warde-Farley, Ozair, Courville, and Bengio]{gans}
Ian Goodfellow, Jean Pouget-Abadie, Mehdi Mirza, Bing Xu, David Warde-Farley,
  Sherjil Ozair, Aaron Courville, and Yoshua Bengio.
\newblock Generative adversarial nets.
\newblock In Z.~Ghahramani, M.~Welling, C.~Cortes, N.~D. Lawrence, and K.~Q.
  Weinberger, editors, \emph{Advances in Neural Information Processing Systems
  27}, pages 2672--2680. Curran Associates, Inc., 2014.
\newblock URL
  \url{http://papers.nips.cc/paper/5423-generative-adversarial-nets.pdf}.

\bibitem[Grathwohl et~al.(2018)Grathwohl, Chen, Bettencourt, Sutskever, and
  Duvenaud]{ffjord}
Will Grathwohl, Ricky~TQ Chen, Jesse Bettencourt, Ilya Sutskever, and David
  Duvenaud.
\newblock Ffjord: Free-form continuous dynamics for scalable reversible
  generative models.
\newblock In \emph{International Conference on Learning Representations}, 2018.

\bibitem[Gulrajani et~al.(2017)Gulrajani, Ahmed, Arjovsky, Dumoulin, and
  Courville]{wgangp}
Ishaan Gulrajani, Faruk Ahmed, Martin Arjovsky, Vincent Dumoulin, and Aaron~C
  Courville.
\newblock Improved training of wasserstein gans.
\newblock In \emph{Advances in neural information processing systems}, pages
  5767--5777, 2017.

\bibitem[{Hao} et~al.(2018){Hao}, {Huang}, and {Belongie}]{flow3}
Z.~{Hao}, X.~{Huang}, and S.~{Belongie}.
\newblock Controllable video generation with sparse trajectories.
\newblock In \emph{2018 IEEE/CVF Conference on Computer Vision and Pattern
  Recognition}, pages 7854--7863, 2018.

\bibitem[He et~al.(2016)He, Zhang, Ren, and Sun]{resnet}
Kaiming He, Xiangyu Zhang, Shaoqing Ren, and Jian Sun.
\newblock Deep residual learning for image recognition.
\newblock In \emph{Proceedings of the IEEE conference on computer vision and
  pattern recognition}, pages 770--778, 2016.

\bibitem[Heusel et~al.(2017)Heusel, Ramsauer, Unterthiner, Nessler, and
  Hochreiter]{ttur}
Martin Heusel, Hubert Ramsauer, Thomas Unterthiner, Bernhard Nessler, and Sepp
  Hochreiter.
\newblock Gans trained by a two time-scale update rule converge to a local nash
  equilibrium.
\newblock In \emph{Advances in neural information processing systems}, pages
  6626--6637, 2017.

\bibitem[Hochreiter and Schmidhuber(1997)]{lstm}
Sepp Hochreiter and J{\"u}rgen Schmidhuber.
\newblock Long short-term memory.
\newblock \emph{Neural computation}, 9\penalty0 (8):\penalty0 1735--1780, 1997.

\bibitem[Hsieh et~al.(2018)Hsieh, Liu, Huang, Fei-Fei, and
  Niebles]{vidrecurrent4}
Jun-Ting Hsieh, Bingbin Liu, De-An Huang, Li~F Fei-Fei, and Juan~Carlos
  Niebles.
\newblock Learning to decompose and disentangle representations for video
  prediction.
\newblock In S.~Bengio, H.~Wallach, H.~Larochelle, K.~Grauman, N.~Cesa-Bianchi,
  and R.~Garnett, editors, \emph{Advances in Neural Information Processing
  Systems 31}, pages 517--526. Curran Associates, Inc., 2018.
\newblock URL
  \url{http://papers.nips.cc/paper/7333-learning-to-decompose-and-disentangle-representations-for-video-prediction.pdf}.

\bibitem[Kahembwe and Ramamoorthy(2019)]{tganf}
Emmanuel Kahembwe and Subramanian Ramamoorthy.
\newblock Lower dimensional kernels for video discriminators.
\newblock \emph{arXiv preprint arXiv:1912.08860}, 2019.

\bibitem[Karras et~al.(2018)Karras, Aila, Laine, and Lehtinen]{proggan}
Tero Karras, Timo Aila, Samuli Laine, and Jaakko Lehtinen.
\newblock Progressive growing of gans for improved quality, stability, and
  variation.
\newblock In \emph{International Conference on Learning Representations}, 2018.

\bibitem[Karras et~al.(2019)Karras, Laine, and Aila]{stylegan1}
Tero Karras, Samuli Laine, and Timo Aila.
\newblock A style-based generator architecture for generative adversarial
  networks.
\newblock In \emph{2019 IEEE/CVF Conference on Computer Vision and Pattern
  Recognition (CVPR)}, pages 4396--4405, 2019.

\bibitem[Karras et~al.(2020)Karras, Laine, Aittala, Hellsten, Lehtinen, and
  Aila]{stylegan2}
Tero Karras, Samuli Laine, Miika Aittala, Janne Hellsten, Jaakko Lehtinen, and
  Timo Aila.
\newblock Analyzing and improving the image quality of stylegan.
\newblock In \emph{Proceedings of the IEEE/CVF Conference on Computer Vision
  and Pattern Recognition}, pages 8110--8119, 2020.

\bibitem[Lee et~al.(2018)Lee, Zhang, Ebert, Abbeel, Finn, and Levine]{savp}
Alex~X. Lee, Richard Zhang, Frederik Ebert, Pieter Abbeel, Chelsea Finn, and
  Sergey Levine.
\newblock Stochastic adversarial video prediction, 2018.

\bibitem[Li et~al.(2020{\natexlab{a}})Li, Yuan, Fang, and Wang]{moflowgan}
Wei Li, Zehuan Yuan, Xiangzhong Fang, and Changhu Wang.
\newblock Moflowgan: Video generation with flow guidance.
\newblock In \emph{2020 IEEE International Conference on Multimedia and Expo
  (ICME)}, pages 1--6. IEEE, 2020{\natexlab{a}}.

\bibitem[Li et~al.(2020{\natexlab{b}})Li, Wong, Chen, and Duvenaud]{sde}
Xuechen Li, Ting-Kam~Leonard Wong, Ricky~TQ Chen, and David Duvenaud.
\newblock Scalable gradients for stochastic differential equations.
\newblock \emph{arXiv preprint arXiv:2001.01328}, 2020{\natexlab{b}}.

\bibitem[Li et~al.(2018)Li, Fang, Yang, Wang, Lu, and Yang]{flow4}
Yijun Li, Chen Fang, Jimei Yang, Zhaowen Wang, Xin Lu, and Ming-Hsuan Yang.
\newblock Flow-grounded spatial-temporal video prediction from still images.
\newblock In \emph{Proceedings of the European Conference on Computer Vision
  (ECCV)}, 9 2018.

\bibitem[Liang et~al.(2017)Liang, Lee, Dai, and Xing]{flow1}
Xiaodan Liang, Lisa Lee, Wei Dai, and Eric~P. Xing.
\newblock Dual motion gan for future-flow embedded video prediction.
\newblock In \emph{Proceedings of the IEEE International Conference on Computer
  Vision (ICCV)}, 10 2017.

\bibitem[Liu et~al.(2017)Liu, Yeh, Tang, Liu, and Agarwala]{flow2}
Ziwei Liu, Raymond~A Yeh, Xiaoou Tang, Yiming Liu, and Aseem Agarwala.
\newblock Video frame synthesis using deep voxel flow.
\newblock In \emph{Proceedings of the IEEE International Conference on Computer
  Vision}, pages 4463--4471, 2017.

\bibitem[Luc et~al.(2020)Luc, Clark, Dieleman, Casas, Doron, Cassirer, and
  Simonyan]{vidrecurrent6}
Pauline Luc, Aidan Clark, Sander Dieleman, Diego de~Las Casas, Yotam Doron,
  Albin Cassirer, and Karen Simonyan.
\newblock Transformation-based adversarial video prediction on large-scale
  data.
\newblock \emph{arXiv preprint arXiv:2003.04035}, 2020.

\bibitem[Miyato et~al.(2018)Miyato, Kataoka, Koyama, and Yoshida]{sngan}
Takeru Miyato, Toshiki Kataoka, Masanori Koyama, and Yuichi Yoshida.
\newblock Spectral normalization for generative adversarial networks.
\newblock In \emph{International Conference on Learning Representations}, 2018.

\bibitem[Qin et~al.(2018)Qin, Mitra, and Wonka]{lossfuncgan}
Yipeng Qin, Niloy Mitra, and Peter Wonka.
\newblock How does lipschitz regularization influence gan training?, 2018.

\bibitem[Radford et~al.(2015)Radford, Metz, and Chintala]{dcgan}
Alec Radford, Luke Metz, and Soumith Chintala.
\newblock Unsupervised representation learning with deep convolutional
  generative adversarial networks.
\newblock \emph{arXiv preprint arXiv:1511.06434}, 2015.

\bibitem[Ranzato et~al.(2014)Ranzato, Szlam, Bruna, Mathieu, Collobert, and
  Chopra]{vidrecurrent1}
MarcAurelio Ranzato, Arthur Szlam, Joan Bruna, Michael Mathieu, Ronan
  Collobert, and Sumit Chopra.
\newblock Video (language) modeling: a baseline for generative models of
  natural videos.
\newblock \emph{arXiv preprint arXiv:1412.6604}, 2014.

\bibitem[Saito et~al.(2017)Saito, Matsumoto, and Saito]{tgan}
Masaki Saito, Eiichi Matsumoto, and Shunta Saito.
\newblock Temporal generative adversarial nets with singular value clipping.
\newblock In \emph{Proceedings of the IEEE international conference on computer
  vision}, pages 2830--2839, 2017.

\bibitem[Saito et~al.(2020)Saito, Saito, Koyama, and Kobayashi]{tganv2}
Masaki Saito, Shunta Saito, Masanori Koyama, and Sosuke Kobayashi.
\newblock Train sparsely, generate densely: Memory-efficient unsupervised
  training of high-resolution temporal gan.
\newblock \emph{International Journal of Computer Vision}, May 2020.
\newblock \doi{10.1007/s11263-020-01333-y}.
\newblock URL \url{https://doi.org/10.1007/s11263-020-01333-y}.

\bibitem[Salimans et~al.(2016)Salimans, Goodfellow, Zaremba, Cheung, Radford,
  and Chen]{inceptionscore}
Tim Salimans, Ian Goodfellow, Wojciech Zaremba, Vicki Cheung, Alec Radford, and
  Xi~Chen.
\newblock Improved techniques for training gans.
\newblock In \emph{Advances in neural information processing systems}, pages
  2234--2242, 2016.

\bibitem[Soomro et~al.(2012)Soomro, Zamir, and Shah]{ucf101}
Khurram Soomro, Amir~Roshan Zamir, and Mubarak Shah.
\newblock Ucf101: A dataset of 101 human actions classes from videos in the
  wild.
\newblock \emph{arXiv preprint arXiv:1212.0402}, 2012.

\bibitem[Srivastava et~al.(2015)Srivastava, Mansimov, and
  Salakhudinov]{vidrecurrent2}
Nitish Srivastava, Elman Mansimov, and Ruslan Salakhudinov.
\newblock Unsupervised learning of video representations using lstms.
\newblock In \emph{International conference on machine learning}, pages
  843--852, 2015.

\bibitem[Tran et~al.(2015)Tran, Bourdev, Fergus, Torresani, and Paluri]{c3d}
Du~Tran, Lubomir Bourdev, Rob Fergus, Lorenzo Torresani, and Manohar Paluri.
\newblock Learning spatiotemporal features with 3d convolutional networks.
\newblock In \emph{Proceedings of the IEEE international conference on computer
  vision}, pages 4489--4497, 2015.

\bibitem[Tulyakov et~al.(2018)Tulyakov, Liu, Yang, and Kautz]{mocogan}
Sergey Tulyakov, Ming-Yu Liu, Xiaodong Yang, and Jan Kautz.
\newblock Mocogan: Decomposing motion and content for video generation.
\newblock In \emph{Proceedings of the IEEE conference on computer vision and
  pattern recognition}, pages 1526--1535, 2018.

\bibitem[Tzen and Raginsky(2019)]{sde2}
Belinda Tzen and Maxim Raginsky.
\newblock Neural stochastic differential equations: Deep latent gaussian models
  in the diffusion limit.
\newblock \emph{arXiv preprint arXiv:1905.09883}, 2019.

\bibitem[Villegas et~al.(2019)Villegas, Pathak, Kannan, Erhan, Le, and
  Lee]{stochhighfid}
Ruben Villegas, Arkanath Pathak, Harini Kannan, Dumitru Erhan, Quoc~V Le, and
  Honglak Lee.
\newblock High fidelity video prediction with large stochastic recurrent neural
  networks.
\newblock In \emph{Advances in Neural Information Processing Systems}, pages
  81--91, 2019.

\bibitem[Vondrick et~al.(2016)Vondrick, Pirsiavash, and Torralba]{vgan}
Carl Vondrick, Hamed Pirsiavash, and Antonio Torralba.
\newblock Generating videos with scene dynamics.
\newblock In \emph{Advances in neural information processing systems}, pages
  613--621, 2016.

\bibitem[Xingjian et~al.(2015)Xingjian, Chen, Wang, Yeung, Wong, and
  Woo]{clstm}
SHI Xingjian, Zhourong Chen, Hao Wang, Dit-Yan Yeung, Wai-Kin Wong, and
  Wang-chun Woo.
\newblock Convolutional lstm network: A machine learning approach for
  precipitation nowcasting.
\newblock In \emph{Advances in neural information processing systems}, pages
  802--810, 2015.

\bibitem[Yildiz et~al.(2019)Yildiz, Heinonen, and Lahdesmaki]{ode2vae}
Cagatay Yildiz, Markus Heinonen, and Harri Lahdesmaki.
\newblock Ode2vae: Deep generative second order odes with bayesian neural
  networks.
\newblock In H.~Wallach, H.~Larochelle, A.~Beygelzimer, F.~d\textquotesingle
  Alch\'{e}-Buc, E.~Fox, and R.~Garnett, editors, \emph{Advances in Neural
  Information Processing Systems 32}, pages 13412--13421. Curran Associates,
  Inc., 2019.
\newblock URL
  \url{http://papers.nips.cc/paper/9497-ode2vae-deep-generative-second-order-odes-with-bayesian-neural-networks.pdf}.

\end{thebibliography}

\end{document}